\documentclass[conference]{IEEEtran}
\IEEEoverridecommandlockouts

\usepackage{cite}
\usepackage{amsmath,amssymb,amsfonts}
\usepackage{algorithm}
\usepackage{algorithmic}
\usepackage{graphicx}
\usepackage{textcomp}
\usepackage{xcolor}
\usepackage{setspace} 

\usepackage{booktabs} 
\usepackage{multirow} 
\usepackage{array} 
\usepackage{mathtools}

\usepackage[colorlinks,linkcolor=red,anchorcolor=blue,citecolor=green]{hyperref} 
\def\BibTeX{{\rm B\kern-.05em{\sc i\kern-.025em b}\kern-.08em
    T\kern-.1667em\lower.7ex\hbox{E}\kern-.125emX}}
\begin{document}

\newcommand{\firstpara}[1]{\noindent\textbf{{#1}.}~~} 

\title{EOOD: Entropy-based Out-of-distribution Detection\\
}

\author{
\IEEEauthorblockN{
		Guide Yang\IEEEauthorrefmark{2}\textsuperscript{1,2}, 
		Chao Hou\IEEEauthorrefmark{2}\textsuperscript{1,2},
        Weilong Peng\textsuperscript{2,3},
        Xiang Fang\textsuperscript{4},
        Yongwei Nie\textsuperscript{5},
        Peican Zhu\textsuperscript{6},
		and Keke Tang\IEEEauthorrefmark{1}\textsuperscript{1,2} \thanks{\IEEEauthorrefmark{2}Joint first authors. \IEEEauthorrefmark{1}Corresponding author.}}
        
\IEEEauthorblockA{\textsuperscript{1}{\textit{Cyberspace Institute of Advanced Technology, Guangzhou University, Guangzhou, China}} \\
}

\IEEEauthorblockA{\textsuperscript{2}{\textit{Huangpu Research School, Guangzhou University, Guangzhou, China}} \\
}

\IEEEauthorblockA{\textsuperscript{3}{\textit{School of Computer Science and Cyber Engineering, Guangzhou University, Guangzhou, China}} \\
}
\IEEEauthorblockA{\textsuperscript{4}{\textit{IGP-ERI@N, Nanyang Technological University, Singapore, Singapore}} \\
}
\IEEEauthorblockA{\textsuperscript{5}{\textit{School of Computer Science \& Engineering, South China University of Technology, Guangzhou, China}} \\
}
\IEEEauthorblockA{\textsuperscript{6}{\textit{School of Artificial Intelligence, OPtics and ElectroNics (iOPEN), Northwestern Polytechnical University, Xi'an, China}} \\
}
\IEEEauthorblockA{yanggidear@gmail.com, houchaohk@gmail.com, wlpeng@tju.edu.cn, xfang9508@gmail.com, nieyongwei@scut.edu.cn, }
\IEEEauthorblockA{ericcan@nwpu.edu.cn, tangbohutbh@gmail.com}
}
\maketitle

\begin{abstract}
Deep neural networks (DNNs) often exhibit overconfidence when encountering out-of-distribution (OOD) samples, posing significant challenges for deployment. Since DNNs are trained on in-distribution (ID) datasets, the information flow of ID samples through DNNs inevitably differs from that of OOD samples. In this paper, we propose an Entropy-based Out-Of-distribution Detection (EOOD) framework. EOOD first identifies specific block where the information flow differences between ID and OOD samples are more pronounced, using both ID and pseudo-OOD samples. It then calculates the conditional entropy on the selected block as the OOD confidence score. Comprehensive experiments conducted across various ID and OOD settings demonstrate the effectiveness of EOOD in OOD detection and its superiority over state-of-the-art methods. 


\end{abstract}

\begin{IEEEkeywords}
Entropy, Out-of-distribution Detection, Deep Neural Networks.
\end{IEEEkeywords}

\section{Introduction}
Deep neural networks (DNNs)~\cite{Lecun-2015-DeepLearning,tang2024reparameterization,he-2016-deep-resnet,tang-DFN,2023RepPVConv} are widely used across various fields and have achieved significant success in high-risk applications such as autonomous driving~\cite{filos2020can,janai2020computer} and medical diagnosis assistance~\cite{esteva2021deep,pooch2020can}. Existing research indicates that neural networks can assign high-confidence predictions to test inputs that do not belong to any of the training classes~\cite{Goodfellow-2014-FGSM,nguyen2015deep,tang2023ood,hein2019relu,he2023deep,10393853}, we refer to such samples as out-of-distribution~(OOD)~samples. If the system fails to recognize these inputs as unknown and alert the user, dangerous consequences might be incurred. Therefore, OOD detection~\cite{tang2023matching,tang2025SimOOD,qiao2023advscod,fang2024your} is crucial for ensuring the safety and reliability of neural network deployments.

One popular method for OOD detection is the Outlier Exposure~(OE)~\cite{hendrycks2018deep} approach, which involves using an auxiliary OOD dataset during the neural network's training epochs to help the model learn the distinction between in-distribution (ID) and OOD samples. Methods like OE~\cite{hendrycks2018deep}, MixOE~\cite{zhang2023mixture}, and DivOE~\cite{zhu2024diversified} have shown promising results. However, utilizing auxiliary OOD datasets may reduce classification accuracy and increase training costs. Moreover, in a truly open-world environment, the distribution of test data may not align with the auxiliary OOD dataset, which might not always be practical in real-world scenarios. Post-hoc OOD detection algorithms significantly alleviate the aforementioned limitations because they do not require retraining or auxiliary OOD datasets, offering advantages in practical applications.

\begin{figure}[t]
\centering
\includegraphics[width=1.0\columnwidth]{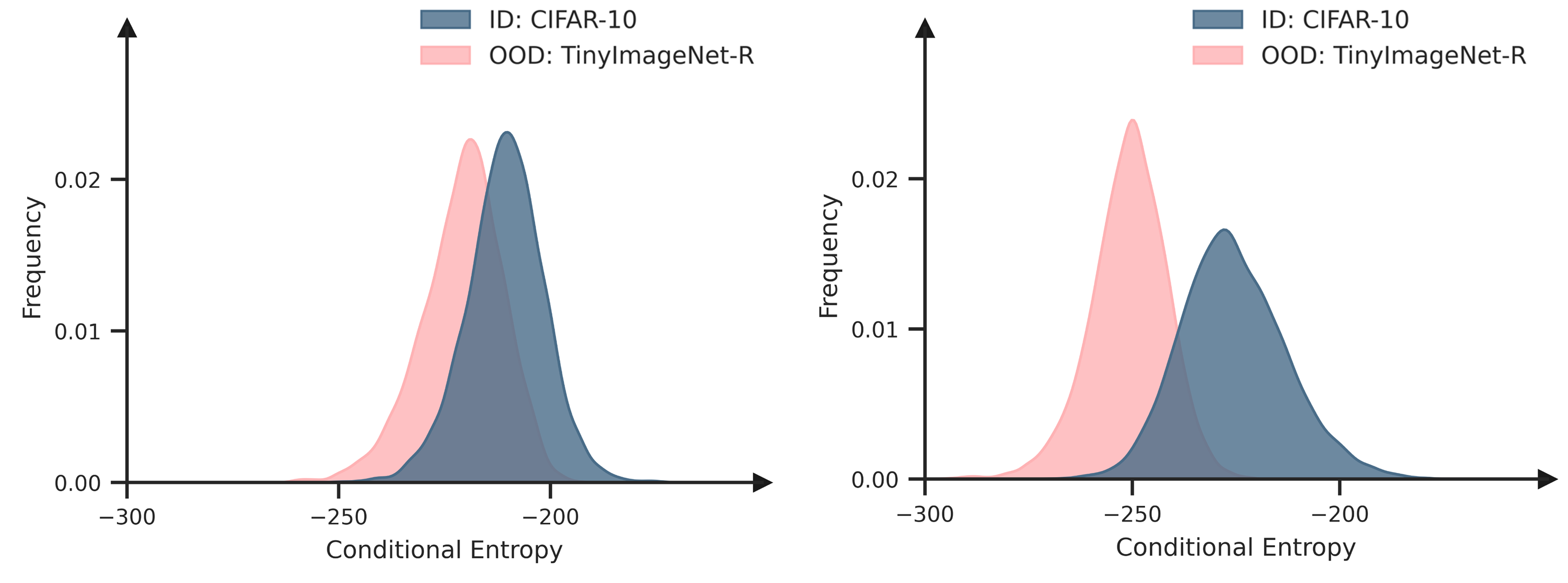}
\vspace{-2mm}
\caption{Visualization of conditional entropy values on WideResNet-28 model. The CIFAR-10 dataset is used as ID, and the TinyImageNet-R dataset is used as OOD. \textbf{Left}: The conditional entropy distribution of the last Block. \textbf{Right}: The conditional entropy distribution of the $10$-th Block selected using the conditional entropy ratio. The $x$-axis represents the conditional entropy computed by $f^{CE}(\cdot, \cdot)$ from Eq.~\ref{eq5}, and the $y$-axis represents the frequency of occurrence. }
\label{fig:fig1}
\end{figure}

In practice, various outputs from neural networks can serve as indicators for distinguishing between in-distribution (ID) and out-of-distribution (OOD) samples. These include the maximum probability from the softmax layer~\cite{hendrycks2016baseline}, its enhancements~\cite{liu2020energy}, and feature values from the penultimate layer~\cite{sun2021react,zhu2022boosting}. However, these signals, primarily designed for ID classification, may not be suitable for OOD detection. Therefore, we analyze the differential behavior of DNNs when encountering OOD samples from the perspective of information flow. Since DNNs are trained on ID datasets, the information flow of ID samples through DNNs inevitably differs from that of OOD samples. We use conditional entropy to capture these differences as a reliable signal for distinguishing between ID and OOD samples. As depicted in Fig.~\ref{fig:fig1}~(right), within specific blocks of DNNs, there is a clear difference in the conditional entropy distributions between ID and OOD samples, demonstrating that conditional entropy effectively captures the differences in information flow between ID and OOD, thereby distinguishing the two. Moreover, by comparing the two subfigures in Fig.~\ref{fig:fig1}, it is evident that the conditional entropy distributions vary across different blocks after the samples pass through them.

In this paper, we propose an \textbf{E}ntropy-based \textbf{O}ut-\textbf{O}f-distribution \textbf{D}etection (\textbf{EOOD}) framework, which is a novel OOD detection method that does not rely on signals related to ID classification. The framework consists of: (1) \emph{\textbf{Selecting Sensitive Block}}: without using auxiliary OOD datasets, we select the most suitable block for OOD detection using our proposed conditional entropy ratio. (2) \emph{\textbf{Calculating the EOOD score}}: based on the selected more sensitive block, we compute its conditional entropy as the final score. The detailed proposed conditional entropy-based OOD detection framework is illustrated in Fig.~\ref{fig:fig2}. We demonstrate the effectiveness of EOOD across multiple neural network architectures. Extensive experiments show the robustness of EOOD in OOD detection and its performance advantages compared to existing methods.

\begin{itemize}
\item[$\bullet$] We propose a simple and effective post-hoc OOD detection method called EOOD, which uses conditional entropy to measure the information change in samples passing through neural networks, thereby further distinguishing ID from OOD. To our knowledge, we are the first to explore the use of conditional entropy in post-hoc OOD detection.

\item[$\bullet$] We calculate the conditional entropy ratio using ID and \emph{pseudo}-OOD, fully considering the most suitable block for OOD detection, thereby achieving better OOD detection performance.

\item[$\bullet$] We extensively evaluate our framework on common benchmarks and establish state-of-the-art performance in post-hoc OOD detection algorithms.
\end{itemize}




\section{Related Work}
\subsection{OOD Detection}\label{AA} 
Since DNNs operate under the closed-world assumption~\cite{krizhevsky2017imagenet}, they face the issue of OOD overconfidence, whereby OOD samples are given high-confidence predictions similar to those of ID samples~\cite{nguyen2015deep}. This presents a significant threat to the deployment of DNNs in open-world scenarios. To address the issue of overconfidence in OOD predictions during the DNN training phase, many methods have been explored. 
Hendrycks \emph{et al.}~\cite{hendrycks2018deep} proposed Outlier Exposure~(OE), which utilizes auxiliary OOD datasets to train the model to have lower confidence on these auxiliary OOD datasets, thereby better calibrating the network. Recent algorithms, DOE~\cite{wang2023out} and MixOE~\cite{zhang2023mixture}, have further extended OE. Additionally, Tang \emph{et al.}~\cite{tang2021codes} used advanced generative models, such as Chamfer GAN, to generate OOD data for training. Meanwhile, Du~\emph{et al.}~\cite{du2022vos} proposed the VOS, while Tao \emph{et al.}~\cite{tao2023nonparametric} proposed the NPOS, both of which synthesize virtual OOD data at a non-image level for training rather than generating specific OOD image samples, thereby further enhancing the generalizability of OE. Although these methods have achieved excellent performance, they all require retraining the network and often necessitate the use of auxiliary OOD datasets. The use of auxiliary OOD datasets can lead to a decrease in accuracy and has several limitations.



\begin{figure*}[t] 
\centering
\includegraphics[width=\textwidth]{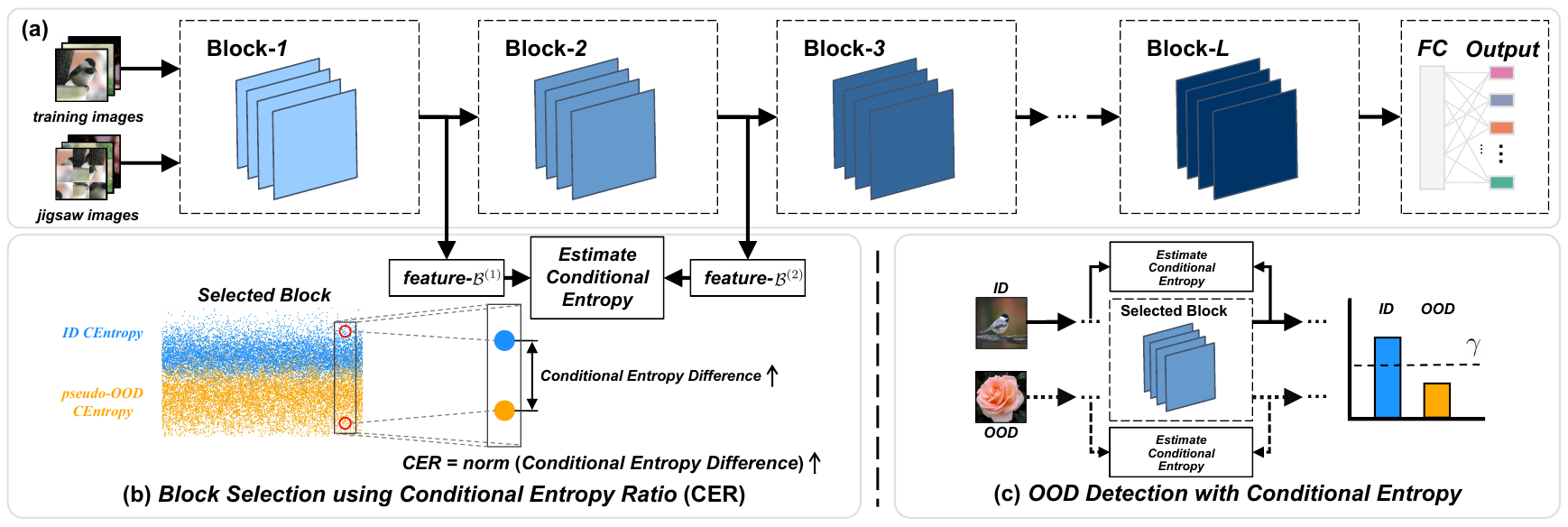}
\caption{Illustration of our proposed out-of-distribution detection framework. (a) Inference path in the neural network, where the input passes through blocks and fully connected (FC) layers to reach the final classification. (b) Select the block with the largest conditional entropy ratio. (c) Calculate the conditional entropy of the selected block as $\text{Score}_{OOD}$.}
\label{fig:fig2}
\end{figure*}

\subsection{Post-hoc OOD Detection} 
Another main focus of current research is on post-hoc OOD detection methods. Hendrycks \emph{et al.}~\cite{hendrycks2016baseline} proposed the classic MSP (maximum softmax probability) algorithm, which posits that OOD data will generally exhibit lower prediction confidence compared to ID data. Liang \emph{et al.}~\cite{liang2018enhancing} extended MSP with their ODIN~(Out-of-DIstribution detector for Neural networks) method, using temperature scaling to further widen the gap between ID and OOD data. Subsequently, Liu \emph{et al.}~\cite{liu2020energy} introduced the Energy Score to reduce prediction bias in energy-based models. Sun \emph{et al.}~\cite{sun2022dice} proposed the DICE~(Directed Sparisification) method, which selected the most important neurons based on their contributions, thereby reducing the impact of OOD data. They also introduced the ReAct~\cite{sun2021react} (Rectified Activations) method, which refines output features through activation correction to better distinguish ID from OOD data. Yuan \emph{et al.}~\cite{yuan2023lhact} improved upon ReAct by proposing Low and High Activations (LHAct), which corrected extremely low activations and employed a new Constrained Butterworth Filter (CBF) to rectify extremely high activations. Tang \emph{et al.}~\cite{tang2024cores} began at the fully connected layers of DNNs, traced back from extreme predictions to identify sample-relevant kernels, and then designed OOD scores based on the responses from these kernels to differentiate between ID and OOD. Zhu \emph{et al.}~\cite{zhu2022boosting} suggested correcting features to a typical set (TFEM) to suppress extreme features, making their approach compatible with various existing OOD algorithms. He~\emph{et al.}~\cite{he2024exploring} built upon TFEM by leveraging the phenomenon that different channels contributed differently to OOD detection and proposed the expLoring channel-Aware tyPical featureS (LAPS) method to enhance OOD detection.


However, most of these scores are derived from ID classification tasks and are not specifically designed for OOD detection. As a result, they may face issues in OOD detection tasks and may not provide the optimal solution. Our method distinguishes between ID and OOD by monitoring the changes in information as samples pass through the neural network, specifically employing conditional entropy to capture variations in the amount of information. This method is inherently independent of the ID classification task and is less affected by the network's classification accuracy. It can focus on the essential ID features, thereby better addressing the OOD detection task. Our method also falls within the category of post-hoc methods, and we compare it with other post-hoc methods discussed in this paper. 

\subsection{Entropy} 
Entropy helps us understand the information content of features. In the realm of information theory, Naftali Tishby \emph{et al.}~\cite{tishby2000information} proposed the information bottleneck theory of deep learning, which posits that DNNs compress inputs as much as possible without sacrificing the ability to accurately predict labels.

Existing works have used entropy for OOD detection. For example, MaxEnt Loss~\cite{neo2024maxent}, based on the maximum entropy principle, proposed a new loss function to address the OOD shift problem. Learning by Erasing~\cite{xing2024learning} utilized entropy to propose a transferable OOD detection method based on deep generative models (DGM). However, while the aforementioned methods have utilized entropy in OOD detection, they require DNNs to be retrained. Our approach applies entropy in a post-hoc manner for OOD detection, thereby eliminating the need for retraining DNNs and significantly reducing the deployment time in real-world environments.

\section{Preliminaries}
This work considers the problem of multi-class classification, where the model is trained using the dataset $D_{in}=\{ (x_i,y_i)\}_{i=1}^I$, with training samples $x_i\in \mathbb{R}^{3\times W\times H}$ being RGB images and $y_i\in\{1,2,\ldots K\}$ being the corresponding labels, where $K$  represents the number of classes. During inference, given an input image sample $x$, the DNN classifier $f$, composed of $L$ blocks, performs inference in the following manner:
\begin{equation}
f:x\rightarrow \mathcal{B}^{(1)}\rightarrow \mathcal{B}^{(2)}\rightarrow\ldots\rightarrow \mathcal{B}^{(L)}\rightarrow \mathcal{O}
\end{equation}
where $\mathcal{B}^{(l)}$ represents the feature map at the $l$-th block in the neural network, and $\mathcal{O}$  represents the final output. The classification is done by identifying which category in $\mathcal{O}$ has the highest predictive confidence.
\begin{equation}
c_{max}=\arg\max \mathcal{O}_i
\end{equation}

The classifier $f$ predicts that the input $x$ belongs to the class $c_{max}$. However, upon deploying $f$ in an open-world setting, we will inevitably encounter OOD samples. In such cases, $f$ outputs a high confidence level for OOD samples, which is contrary to our expectations.

A strategy for OOD detection involves constructing a score, i.e., $\text {Score}_{OOD}$, which distinguishes between ID and OOD through a threshold~\cite{hendrycks2016baseline}.

\begin{equation}
g_\gamma(\mathbf{x})= 
\begin{cases}
\text {ID} & \text {Score}_{OOD} \geq \gamma \\ 
\text {OOD} & \text {Score}_{OOD}<\gamma\end{cases}
\end{equation}
where $\gamma$ is the threshold, which is typically chosen to correctly classify 95\% of the ID data.

\begin{table*}[t]
\centering
\caption{Comparison on the OOD detection performance of different methods under the small-scale setting. The best and second-best results are respectively highlighted with \textbf{bold} and \underline{underlined}.}
\setlength{\tabcolsep}{2pt} 
\resizebox{1.0\textwidth}{!}{
\setstretch{1.4}{
\begin{tabular}{c|c|c|cccccccccccccc}
\bottomrule 
\multirow{3}{*}{ID}         & \multirow{3}{*}{Model}         & \multirow{3}{*}{Method} & \multicolumn{12}{c}{OOD}                                                                                                                                                                                      & \multicolumn{2}{c}{\multirow{2}{*}{Average}} \\ \cline{4-15}
                            &                                &                         & \multicolumn{2}{c}{Textures}    & \multicolumn{2}{c}{SVHN}        & \multicolumn{2}{c}{iSUN}        & \multicolumn{2}{c}{LSUN-R} & \multicolumn{2}{c}{TinyImageNet-R} & \multicolumn{2}{c}{LSUN}        & \multicolumn{2}{c}{}                         \\ \cline{4-17} 
                            &                                &                         & FPR95         & AUROC         & FPR95         & AUROC         & FPR95         & AUROC         & FPR95          & AUROC         & FPR95           & AUROC          & FPR95         & AUROC         & FPR95↓                & AUROC↑               \\ \hline
\multirow{14}{*}{\rotatebox{90}{CIFAR-10}}  & \multirow{7}{*}{\rotatebox{90}{VGG-11}}         & MSP                     & 63.86          & 89.37          & 68.07          & 90.02          & 71.81          & 85.71          & 70.19           & 86.29          & 74.34            & 83.96           & 46.63          & 93.73          & 65.82                 & 88.18                \\
                            &                                & React                   & 51.73          & 87.47          & 58.81          & 83.28          & 51.30          & 88.07          & 47.19           & 89.68          & 61.16            & 84.12           & 23.40          & 94.77          & 48.93                 & 87.90                \\
                            &                                & Energy                  & \underline{47.04}    & \textbf{92.08} & \underline{53.13}    & \underline{92.26}    & 55.39          & 88.97          & 53.02           & 89.58          & 60.29            & 87.32           & 18.51          & 97.20          & 47.90                 & 91.24                \\
                            &                                & ODIN                    & 48.09          & \underline{91.94}    & 53.84          & 92.23          & 56.61          & 88.87          & 54.29           & 89.47          & \underline{54.77}      & \underline{88.36}     & 19.95          & 97.01          & 47.93                 & 91.31                \\
                            &                                & TFEM                    & 49.34          & 91.24          & 61.01          & 90.54          & \underline{46.60}    & \underline{91.11}    & \underline{43.74}     & 91.86          & 60.29            & 87.31           & \underline{16.77}    & \underline{97.28}    & \underline{46.29}           & \underline{91.56}          \\
                            &                                & LAPS                    & 49.41          & 91.18          & 61.10          & 90.46          & 46.70          & 91.10          & 43.79           & \underline{91.87}    & 60.35            & 87.29           & 16.85          & 97.27          & 46.37                 & 91.53                \\
                            &                                & EOOD                 & \textbf{34.47} & 90.89          & \textbf{22.58} & \textbf{95.08} & \textbf{21.54} & \textbf{95.80} & \textbf{25.50}  & \textbf{94.89} & \textbf{26.60}   & \textbf{94.51}  & \textbf{11.59} & \textbf{97.64} & \textbf{23.71}        & \textbf{94.80}       \\ \cline{2-17} 
                            & \multirow{7}{*}{\rotatebox{90}{WideResNet-28}} & MSP                     & 53.30          & 87.45          & 42.10          & 91.85          & 40.11          & 93.05          & 37.81           & 93.71          & 42.51            & 92.12           & 22.70          & 96.69          & 39.76                 & 92.48                \\
                            &                                & React                   & 44.38          & 91.97          & \underline{15.92}    & \textbf{97.06} & 31.52          & 94.17          & 27.98           & 94.75          & 35.64            & \underline{93.76}     & 43.24          & 92.75          & 33.11                 & 94.08                \\
                            &                                & Energy                  & 46.06          & 85.09          & 33.11          & 90.54          & 25.12          & 94.17          & \underline{22.68}     & 94.90          & 32.29            & 92.88           & \underline{9.98}     & \underline{98.36}    & 28.21                 & 92.66                \\
                            &                                & ODIN                    & 47.58          & 82.85          & 37.08          & 88.36          & \textbf{22.95} & 94.22          & \textbf{20.51}  & 95.03          & \underline{28.60}      & 90.24           & 11.63          & 97.52          & \underline{28.06}           & 91.37                \\
                            &                                & TFEM                    & 45.30          & 91.18          & 50.60          & 89.50          & 26.14          & \underline{94.70}    & 22.94           & \textbf{95.56} & 32.29            & 92.88           & \textbf{9.98}  & \textbf{98.36} & 31.21                 & \underline{93.70}          \\
                            &                                & LAPS                    & \underline{38.63}    & \underline{92.68}    & 24.38          & 94.87          & 34.97          & 91.54          & 32.71           & 91.66          & 35.65            & 92.32           & 13.36          & 96.95          & 29.95                 & 93.34                \\
                            &                                & EOOD                 & \textbf{22.71} & \textbf{93.76} & \textbf{11.68} & \underline{96.92}    & \underline{24.63}    & \textbf{96.05} & 32.75           & \underline{95.05}    & \textbf{26.74}   & \textbf{95.50}  & 12.37          & 97.88          & \textbf{21.81}        & \textbf{95.86}       \\ \hline
\multirow{14}{*}{\rotatebox{90}{CIFAR-100}} & \multirow{7}{*}{\rotatebox{90}{VGG-16}}         & MSP                     & 87.48          & 68.63          & 85.37          & 73.85          & 83.61          & 74.37          & 79.42           & 78.07          & 82.42            & 73.89           & 72.49          & 78.82          & 81.80                 & 74.61                \\
                            &                                & React                   & 82.82          & \underline{74.97}    & 86.13          & 73.51          & 74.49          & 79.43          & 68.52           & 83.16          & 72.74            & \underline{79.02}     & 58.23          & 83.67          & 73.82                 & 78.96                \\
                            &                                & Energy                  & 84.11          & 70.39          & 82.62          & 75.72          & 77.66          & 77.19          & 70.21           & 81.73          & 74.98            & 77.11           & \underline{54.40}    & \underline{83.73}    & 74.00                 & 77.65                \\
                            &                                & ODIN                    & 84.22          & 66.67          & 82.94          & 71.58          & \underline{73.32}    & 76.92          & \underline{66.10}     & 81.28          & \underline{71.06}      & 76.93           & 57.01          & 81.46          & \underline{72.44}           & 75.81                \\
                            &                                & TFEM                    & 83.78          & 70.64          & 79.34          & 79.91          & 75.07          & \underline{79.72}    & 69.33           & \underline{83.27}    & 72.97            & 78.99           & 60.69          & 82.15          & 73.53                 & 79.11                \\
                            &                                & LAPS                    & \underline{81.81}    & \textbf{78.18} & \underline{70.53}    & \underline{84.48}    & 82.21          & 76.59          & 75.77           & 79.00          & 80.90            & 73.61           & \textbf{51.37} & \textbf{90.46} & 73.77                 & \underline{80.39}          \\
                            &                                & EOOD                 & \textbf{68.16} & 70.30          & \textbf{15.94} & \textbf{95.17} & \textbf{35.74} & \textbf{89.77} & \textbf{22.58}  & \textbf{94.68} & \textbf{39.05}   & \textbf{88.73}  & 78.92          & 68.42          & \textbf{43.40}        & \textbf{84.51}       \\ \cline{2-17} 
                            & \multirow{7}{*}{\rotatebox{90}{WideResNet-28}} & MSP                     & 81.13          & 78.25          & 75.31          & 81.68          & 80.02          & 75.54          & 79.85           & 76.69          & 80.47            & 73.65           & 71.65          & 84.09          & 78.07                 & 78.32                \\
                            &                                & React                   & \underline{68.51}    & \textbf{83.00} & 66.05          & 86.61    & 72.92          & 73.77          & 73.27           & 74.45          & 75.94            & 70.91           & 62.44          & 86.29          & 69.86                 & 79.17                \\
                            &                                & Energy                  & 80.11          & 79.04          & 70.93          & 82.94          & 75.56          & 78.02          & 74.42           & 79.35          & 76.82            & 75.72           & 67.82          & 85.33          & 74.28                 & 80.07                \\
                            &                                & ODIN                    & 74.29          & 76.66          & \textbf{27.39} & \textbf{94.39} & \underline{66.36}    & 81.26          & 65.74           & 81.62          & \underline{67.08}      & \underline{79.47}     & \underline{52.40}    & \underline{90.61}    & \underline{58.88}           & \underline{84.00}          \\
                            &                                & TFEM                    & 80.46          & 78.42          & 71.01          & \underline{87.50}          & 72.78          & \underline{81.48}    & \underline{57.01}     & \textbf{88.04} & 76.03            & 78.14           & 72.03          & 81.11          & 71.55                 & 82.45                \\
                            &                                & LAPS                    & 75.82          & \underline{82.76}    & 69.34          & 86.48          & 73.62          & 75.15          & 73.86           & 75.80          & 76.26            & 72.77           & 61.95          & 86.62          & 71.81                 & 79.93                \\
                            &                                & EOOD                 & \textbf{57.16} & 78.60          & \underline{48.76}    & 84.91          & \textbf{54.03} & \textbf{87.16} & \textbf{54.20}  & \underline{87.94}    & \textbf{51.89}   & \textbf{87.39}  & \textbf{7.98}  & \textbf{98.44} & \textbf{45.67}        & \textbf{87.41}       \\ \toprule
\end{tabular}
}
}
\label{tab_1}  
\end{table*}

\section{Method} 
\subsection{Overview of EOOD framework}
\label{sec:framework} 
Our OOD detection framework is based on the idea that the information flow trends of ID and OOD samples differ during the inference process of a pre-trained neural network, with the framework specifically depicted in Fig.~\ref{fig:fig2}. In this section, we first detail how to calculate the conditional entropy of each block (Section \ref{sec:Computing_CE}). Then, we explain how to calculate the conditional entropy ratio using ID and generated \emph{pseudo}-OOD samples to select the most sensitive block for OOD detection (Section \ref{sec:Selecting Sensitive Block}). Finally, we introduce our \textbf{E}ntropy-based \textbf{O}ut-\textbf{O}f-distribution \textbf{D}etection score $\text{Score}_{OOD}$ (Section \ref{sec:Final_Score}).

\subsection{
Block Conditional Entropy}\label{sec:Computing_CE} 
Since DNNs are trained on ID datasets, the information flow of ID samples through DNNs inevitably differs from that of OOD samples. To capture this difference, we compute conditional entropy, which reflects changes in the information flow after images pass through a specific block. A larger conditional entropy indicates greater changes in information flow, and vice versa.

Given an ID sample \(x\) as input, it passes through a DNN, resulting in the output feature map of the \(l\)-th block with \(c\) channels, denoted as: \(\mathcal{B}^{(l)} = \{\mathrm{B}^{(l)}_1, \mathrm{B}^{(l)}_2, \dots, \mathrm{B}^{(l)}_c\}\). We use Shannon entropy~\cite{shannon2001mathematical} to measure the amount of information in $\mathcal{B}^{(l)}$:

\begin{equation}
H(\mathcal{B}^{(l)})=-\int p(\mathrm{B}_i^{(l)})\log p(\mathrm{B}_i^{(l)}) d\mathrm{B}_i^{(l)}
\end{equation}
where $p(\cdot)$ represents the probability density function (PDF). Specifically, considering the computational overhead, we use the NPEET~\cite{steeg2014non} library for estimation, which is based on the $k$-nearest neighbors entropy estimation technique.

From this, for a given sample $x$, we can calculate it's conditional entropy in the \(l\)-th block:
\begin{equation}
\label{eq5}
\begin{aligned}
&f^{CE}(x, l)=H(\mathcal{B}^{(l-1)}|\mathcal{B}^{(l)})=H(\mathcal{B}^{(l-1)},\mathcal{B}^{(l)})-H(\mathcal{B}^{(l)}) \\
\end{aligned}
\end{equation}
where $H(\mathcal{B}^{(l-1)},\mathcal{B}^{(l)})$ is estimated by concatenating the $\mathcal{B}^{(l-1)}$ and $\mathcal{B}^{(l)}$ to obtain the Shannon entropy. 

A lower conditional entropy indicates greater similarity between the feature maps before and after a block, i.e., less information change. As depicted in Fig.~\ref{fig:fig1}, we observe significant differences in the conditional entropy values between ID and OOD data within specific blocks. These blocks demonstrate higher efficiency in information compression and are more sensitive to the flow of ID information, thereby more effectively distinguishing between ID and OOD.

\begin{table*}[]
\caption{Comparison on the OOD detection performance of different methods under the large-scale setting. The best and second-best results are respectively highlighted with \textbf{bold} and \underline{underlined}.}
\centering
\resizebox{1.0\textwidth}{!}{
\setstretch{1.2}{
\begin{tabular}{c|c|c|cccccccccc}
\bottomrule 
\multirow{3}{*}{ID}           & \multirow{3}{*}{Model}         & \multirow{3}{*}{Method} & \multicolumn{8}{c}{OOD}                                                                                                               & \multicolumn{2}{c}{\multirow{2}{*}{Average}} \\ \cline{4-11}
                              &                                &                         & \multicolumn{2}{c}{SUN}         & \multicolumn{2}{c}{PLACES}      & \multicolumn{2}{c}{Textures}    & \multicolumn{2}{c}{ImageNet-O }  & \multicolumn{2}{c}{}                         \\ \cline{4-13} 
                              &                                &                         & FPR95         & AUROC         & FPR95         & AUROC         & FPR95         & AUROC         & FPR95         & AUROC         & FPR95↓                & AUROC↑               \\ \hline
\multirow{14}{*}{\rotatebox{90}{ImageNet-1k}} & \multirow{7}{*}{\rotatebox{90}{VGG-16}}        & MSP                     & 75.66          & 78.31          & 77.89          & 77.60          & 64.84          & 81.66          & 96.90          & 52.29          & 78.82                 & 72.47                \\
                              &                                & React                   & 99.87          & 35.01          & 99.25          & 37.54          & 96.45          & 49.12          & 98.80          & 39.56          & 98.59                 & 40.31                \\
                              &                                & Energy                  & 68.13          & 85.89          & 69.37          & 83.91          & 54.91          & 88.88          & 95.25          & \underline{64.08}    & 71.92                 & 80.69                \\
                              &                                & ODIN                    & 61.31          & 86.51          & 67.33          & 83.87          & 44.57          & \textbf{89.82} & 94.95          & 50.60          & 67.04                 & 77.70                \\
                              &                                & TFEM                    & \textbf{44.03} & \textbf{91.50} & \textbf{48.67} & \textbf{88.84} & \underline{37.94}    & \underline{88.99}    & \underline{90.85}    & 56.77          & \textbf{55.37}        & \textbf{81.53}       \\
                              &                                & LAPS                    & 57.37          & \underline{88.79}    & 62.49          & \underline{85.11}    & 49.42          & 86.48          & 91.95          & \textbf{65.07} & 65.31                 & \underline{81.36}          \\
                              &                                & EOOD                 & \underline{49.08}    & 83.19          & \underline{61.87}    & 77.90          & \textbf{36.45} & 86.49          & \textbf{83.20} & 61.67          & \underline{57.65}           & 77.31                \\ \cline{2-13} 
                              & \multirow{7}{*}{\rotatebox{90}{WideResNet-50}} & MSP                     & 72.51          & 76.66          & 73.38          & 75.15          & 77.52          & 68.74          & 95.55          & 51.41          & 79.74                 & 67.99                \\
                              &                                & React                   & 58.47          & 90.10          & 66.50          & \underline{87.34}    & 84.40          & 66.39          & 95.45          & 45.60          & 76.21                 & 72.36                \\
                              &                                & Energy                  & 61.27          & 86.10          & 65.98          & 84.23          & 59.10          & 84.11          & 93.90          & 60.60          & 70.06                 & 78.76                \\
                              &                                & ODIN                    & 65.61          & 83.46          & 64.89          & 82.82          & 90.89          & 63.15          & 96.25          & 49.31          & 79.41                 & 69.69                \\
                              &                                & TFEM                    & \underline{51.05}    & \underline{90.62}    & \underline{61.04}    & 86.80          & \underline{55.27}    & \underline{86.40}    & \underline{88.40}    & \underline{60.60}    & \underline{63.94}           & \underline{81.11}          \\
                              &                                & LAPS                    & \textbf{39.05} & \textbf{93.07} & \textbf{48.50} & \textbf{90.58} & 91.42          & 62.36          & 96.65          & 43.93          & 68.91                 & 72.49                \\
                              &                                & EOOD                 & 58.27          & 86.91          & 68.32          & 81.41          & \textbf{28.71} & \textbf{93.57} & \textbf{84.00} & \textbf{69.70} & \textbf{59.83}        & \textbf{82.90}       \\ \toprule
\end{tabular}
}
}
\label{tab_2}
\end{table*}

\subsection{Selection of Sensitive Block}\label{sec:Selecting Sensitive Block}
To determine which block is more effective, we propose the Conditional Entropy Ratio (CER), which identifies the most sensitive block by calculating the separation degree between the conditional entropy distributions of ID and OOD samples.

\firstpara{Generating Pseudo-OOD Samples}
Actually, obtaining OOD samples before deploying the neural network to the open world is often impractical. Therefore, we generate $pseudo$-OOD samples from ID samples using a $jigsaw$ puzzle approach~\cite{noroozi2016unsupervised}, which retain some features of the ID samples but exhibit a semantic shift, making them a challenging type of OOD to detect~\cite{hsu2020generalized}. Given an ID training dataset \(X=\{x_i\}_{i=1:I}\), we generate and shuffle a \(3 \times 3\) $jigsaw$ puzzle to produce the corresponding $pseudo$-OOD dataset, denoted as \(\hat{X}=\{\hat{x}_i\}_{i=1:I}\).

\firstpara{Computing Conditional Entropy Ratio}
The CER of $l$-th block are represented by \(R^{(l)}\), which reflect the separation degree between the conditional entropy distributions of ID and OOD samples,
\begin{equation}
\label{eq6}
\begin{aligned}
R^{(l)}=\frac{1}{I}\sum_{i=1}^{I}\frac{\Vert f^{CE}(x_i, l)-f^{CE}(\hat{x}_i, l) \Vert_1 }{\max\limits_{1 \le i \le I} \Vert f^{CE}(x_i, l)-f^{CE}(\hat{x}_i, l) \Vert_1} 
\end{aligned}
\end{equation}
where the term \(\Vert \cdot \Vert_1\) represents the $L_1$-norm. For the complete steps to compute the CER, refer to Algorithm~\ref{alg:algorithm}.

Consequently, we can calculate the CER for all blocks using \(f^{CE}(\cdot, \cdot)\), resulting in \(\mathcal{R}\):
\begin{equation}
\begin{aligned}
& \mathcal{R}=\{R^{(1)}, R^{(2)}, \dots, R^{(L)}\} 
\end{aligned}
\end{equation}

\begin{algorithm}[tb]
    \caption{Computing CER in $l$-th Block} 
    \label{alg:algorithm}
    \textbf{Input}: Training data $\{x_i, y_i\}_{i=1}^I$ \\
    \textbf{Output}: CER $R^{(l)}$ in $l$-th Block
    \begin{algorithmic}[1] 
        \WHILE{$1 \le i \le I$}  
            \STATE \textbf{Generating Pseudo-OOD Samples:} %
            \STATE $\hat{x}_i \leftarrow  \emph{Jigsaw}\left(x_i\right)$

        \STATE \textbf{Computing Conditional Entropy:}
        \STATE Conditional Entropies: $f^{CE}(x_i, l)$, $f^{CE}(\hat{x}_i, l)$

        \ENDWHILE
    \STATE \textbf{Computing CER:}
    \STATE Based on Eq.~(\ref{eq6}), we compute \(R^{(l)}\)
    \STATE \textbf{return} CER $R^{(l)}$
    \end{algorithmic}
\end{algorithm}

\firstpara{Selecting Sensitive Block}
The larger the CER, the more sensitive the corresponding block becomes. To identify the most sensitive block, we determine the block index \(l^*\) based on the CER \(\mathcal{R}\),

\begin{equation}
\begin{aligned}
& {l^*}=\arg\max(\mathcal{R})\\
\end{aligned}
\end{equation}
this \(l^*\)-th block is utilized to compute the final EOOD score.

\subsection{Final EOOD Score}
\label{sec:Final_Score}
For a given test image \(x\), we calculate the conditional entropy based on the selected block to serve as the final EOOD score:
\begin{equation}
\text {Score}_{OOD}(x) = f^{CE}(x, l^*)
\end{equation}
this score is used to differentiate between ID and OOD samples.

\section{Experimental Results}
\subsection{Experiment Setup}\label{sec:Setup}
\firstpara{Implementation}
When using CIFAR-10/100~\cite{krizhevsky2009learning} as ID datasets, we employ commonly used DNNs: VGG-11~\cite{zagoruyko2016wide}, WideResNet-28~\cite{zagoruyko2016wide} and VGG-16~\cite{simonyan2014very}. WideResNet-28 and VGG-11 were trained for 100 epochs, while VGG-16 was trained for 200 epochs. For ImageNet-1k~\cite{deng2009imagenet} as the ID dataset, we utilize the pre-trained WideResNet-50 and VGG-16 models released on PyTorch~\cite{paszke2019pytorch}. VGG-11, WideResNet-28, VGG-16, and WideResNet-50 have 8, 12, 13, and 16 blocks, respectively. All experiments are implemented in PyTorch and conducted on a machine equipped with a dual-core 2.40GHz CPU, 128GB of RAM, and four NVIDIA RTX 2080Ti GPUs.

\begin{figure}[t]
\centering
\includegraphics[width=1.0\columnwidth]{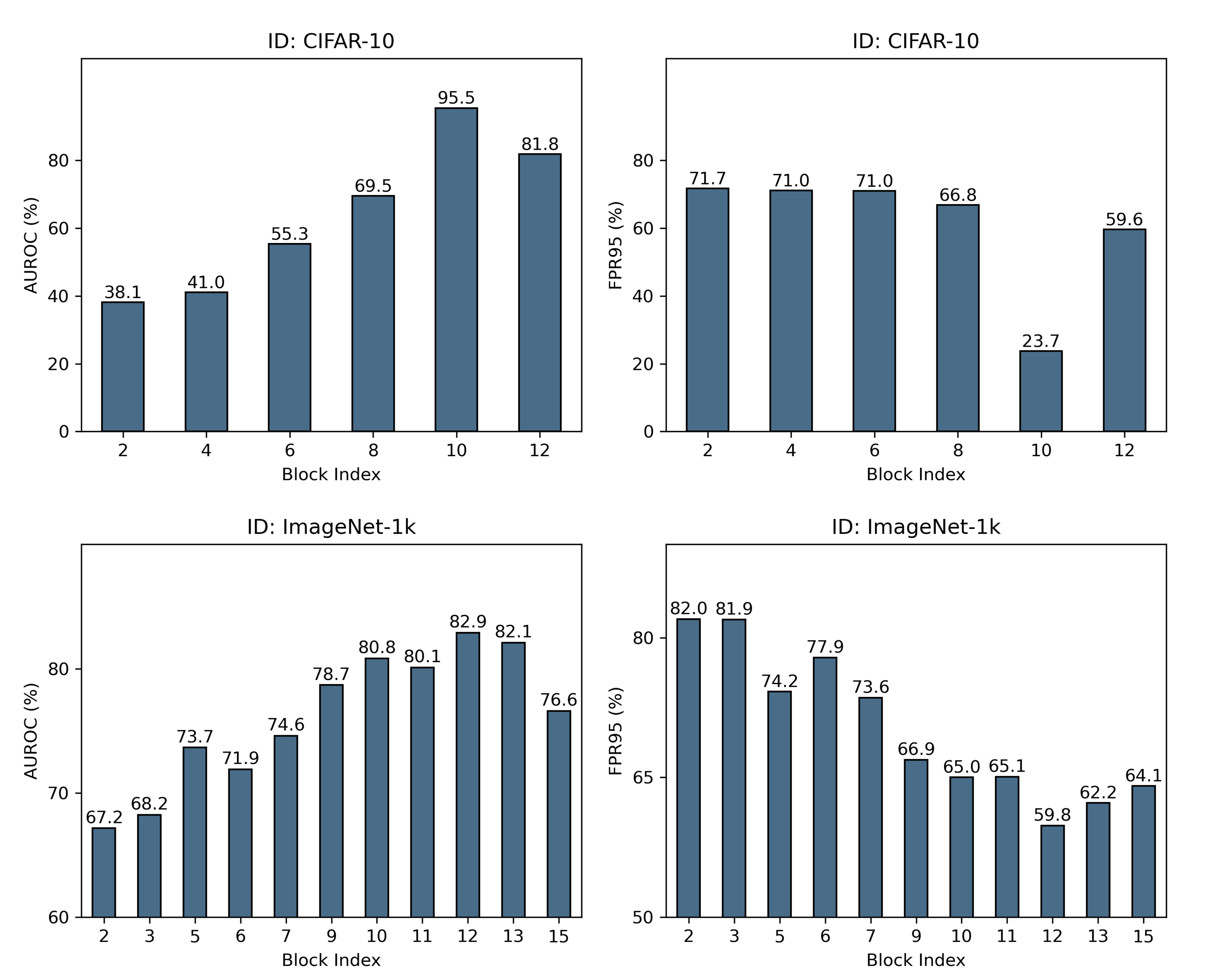}
\caption{Comparison of EOOD's OOD detection performance using different blocks in WideResNet-28 trained on CIFAR-10 and WideResNet-50 trained on ImageNet-1k, with FPR95 and AUROC metrics averaged over multiple OOD datasets.}
\label{fig:fig3}
\end{figure}

\firstpara{ID/OOD Settings}
We investigate two ID/OOD scenarios: (1) Small-scale OOD setting: CIFAR-10 and CIFAR-100 are used as ID datasets, and Textures~\cite{cimpoi2014describing}, SVHN~\cite{netzer2011reading}, iSUN~\cite{xu2015turkergaze}, LSUN~\cite{yu2015lsun}, LSUN-resize (LSUN-R)~\cite{yu2015lsun}, and TinyImageNet-R~\cite{tavanaei2020embedded} are used as the corresponding OOD datasets, with ``-R'' denoting the resized variants. At the same time, TinyImageNet-R is used as \emph{Near-OOD} for CIFAR-10/100 datasets. All images are resized to $32\times32$. (2) Large-scale OOD setting: ImageNet-1k is used as the ID dataset, and SUN~\cite{xiao2010sun}, PLACES~\cite{zhou2017places}, Textures~\cite{cimpoi2014describing}, and ImageNet-O~\cite{hendrycks2021natural} are used as OOD datasets. Notably, ImageNet-O serves as the \emph{Near-OOD} for ImageNet-1k. In particular, Winkens \emph{et al.}~\cite{winkens2020contrastive} categorize OOD into more challenging \emph{Near-OOD} tasks and easier \emph{Far-OOD} tasks. All images are resized to $256\times256$ and then center cropped to a size of $224\times224$.

\firstpara{Baseline Methods}
We compare EOOD against competitive OOD, selecting five post-hoc OOD detection methods as the baseline, including MSP~\cite{hendrycks2016baseline}, ODIN~\cite{liang2018enhancing}, Energy~\cite{liu2020energy}, ReAct~\cite{sun2021react}, TFEM~\cite{zhu2022boosting}, and LAPS~\cite{he2024exploring}. To ensure a fair comparison, all methods use post-hoc algorithms with pre-trained neural networks.

\firstpara{Evaluation Metrics}
We use the two most widely adopted metrics in OOD detection research to measure the quality of OOD detection. \textbf{FPR95}~\cite{liang2018enhancing}: The false positive rate (FPR) of OOD instances when the true positive rate (TPR) of in-distribution instances is as high as 95\%. A lower FPR95 indicates better OOD detection performance, and vice versa.
\textbf{AUROC}~\cite{davis2006relationship}: The area under the receiver operating characteristic curve (ROC), which is composed of the TPR and FPR, is also a threshold-independent measure. A higher AUROC indicates better OOD detection performance, and vice versa. The specific values of FPR95 and AUROC are expressed in percentages for all experiments.

\subsection{Performance Analyses}\label{sec:Results}


\firstpara{Performance on CIFAR benchmarks}
In Tab.~\ref{tab_1}, we report the OOD detection performance of various post-hoc OOD methods on the CIFAR benchmarks across VGG-11, VGG-16, and WideResNet-28 network architectures. We assess performance based on FPR95 and AUROC metrics across six common OOD datasets, with TinyImageNet-R serving as the \emph{Near-OOD} for the CIFAR dataset, which poses a greater challenge for OOD detection. Notably, on the CIFAR-100 dataset, EOOD reduces the average FPR95 by 51.95\% on VGG-16 and 11.57\% on WideResNet-28 compared to the second-best method. These results underscore the effectiveness of Entropy and its significant impact on OOD detection.

\begin{figure}[t]
\centering
\includegraphics[width=0.85\columnwidth]{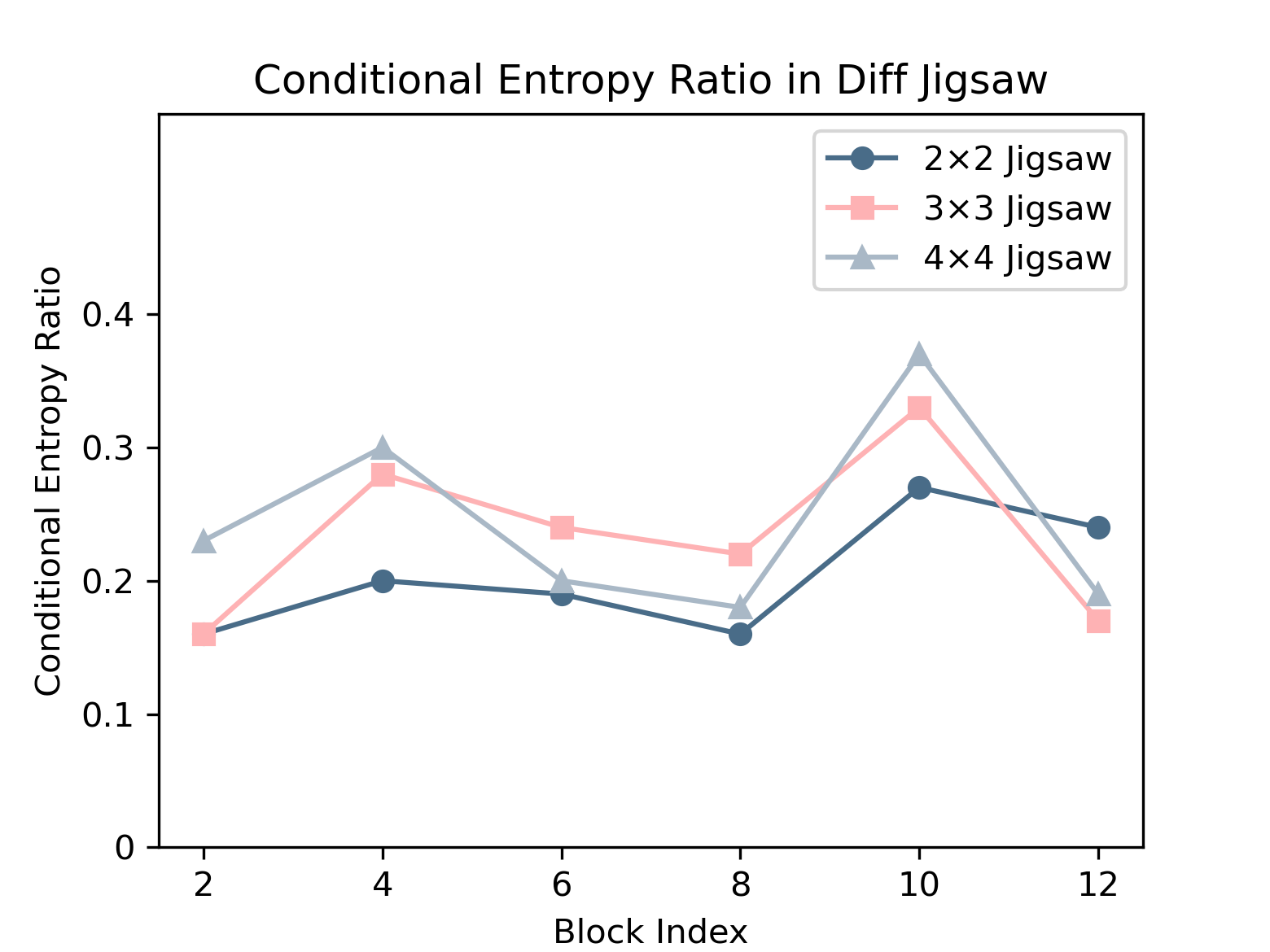} 
\caption{Computing the CER under different jigsaw puzzle configurations using WideResNet-28 model trained on CIFAR-10.}
\label{fig:fig4}
\end{figure}

\firstpara{Performance on ImageNet benchmark}
Subsequently, we report the OOD detection performance of VGG-16 and WideResNet-50 on large-scale datasets in Tab. \ref{tab_2}. This presents a greater challenge, and we calculate the corresponding performance using FPR95 and AUROC on four common OOD datasets. Among these, ImageNet-O serves as the \emph{Near-OOD} dataset for ImageNet-1k, making OOD detection more challenging. Compared to the second-best results obtained with the WideResNet-50 architecture, EOOD reduces the average FPR95 on the ImageNet benchmark by 6.43\% and increases the average AUROC by 2.21\%.

\subsection{Ablation Studies}\label{sec:Ablation Studies}
\firstpara{Effect of Block Selection}
To evaluate the impact of block selection on the performance of EOOD in OOD detection, we conduct experiments using the WideResNet-28 network on the CIFAR-10 dataset and the WideResNet-50 network on the ImageNet-1k dataset. When using CIFAR-10 as ID dataset, we calculate the largest CER to be $0.33$, corresponding to the $10$-th block of the WideResNet-28, which exhibits the best performance among all blocks. Similarly, when using ImageNet-1k as ID dataset, we calculate the largest CER to be $0.08$, corresponding to the $12$-th block of the WideResNet-50, which also exhibits the best performance among all blocks. Additionally, as shown in Fig.~\ref{fig:fig3}, the OOD performance noticeably varies across different blocks. Furthermore, the last block, commonly used for classification, is not the optimal choice for OOD detection. Therefore, it is crucial to select the sensitive block for calibration.

\firstpara{Effect of Different Jigsaw Configurations}
To explore the impact of different jigsaw puzzle settings, we generate $2\times2$, $3\times3$, and $4\times4$ $jigsaw$ puzzles and calculate the CER for each block. Despite the varying sizes of the puzzles, our method consistently identifies the most sensitive block. On the CIFAR-10 benchmark, the $10$-th block of WideResNet-28 consistently demonstrates the best performance. This consistency underscores the effectiveness of our method in selecting the optimal block for OOD detection, as specifically illustrated in Fig.~\ref{fig:fig4}.

\section{Conclusion} 

In this paper, we have proposed a simple and novel entropy-based OOD detection framework. We have used ID and $pseudo$-OOD samples to calculate the \emph{CER} to select more sensitive blocks, then utilized conditional entropy to compute the OOD confidence score for OOD detection. Extensive experiments have validated the effectiveness of conditional entropy in OOD detection and demonstrated its superior performance. Future plans include applying conditional entropy to more security issues, such as adversarial attacks. We hope our research can help address the issue of overconfidence in neural networks and inspire the use of entropy in handling tasks in an open world.

\section*{Acknowledgements}
We sincerely thank the anonymous reviewers for their valuable comments.
This work was supported in part by the National Natural Science Foundation of China (62472117), the Guangdong Basic and Applied Basic Research Foundation (2025A1515010157, 2024A1515012064), 
the Science and Technology Projects in Guangzhou (2025A03J0137, SL2022A04J01112),
and the Academician Binxing Fang's Specialized Class.

\bibliographystyle{IEEEtran} 
\bibliography{mybibliography} 

\vspace{12pt}

\end{document}